\documentclass{article} % For LaTeX2e
\usepackage{iclr2021_conference,times}

\usepackage{amsmath,amsfonts,bm}

\def\eqref#1{equation~\ref{#1}}
\def\1{\bm{1}}

\DeclareMathAlphabet{\mathsfit}{\encodingdefault}{\sfdefault}{m}{sl}
\SetMathAlphabet{\mathsfit}{bold}{\encodingdefault}{\sfdefault}{bx}{n}

\usepackage[utf8]{inputenc} % allow utf-8 input
\usepackage[T1]{fontenc}    % use 8-bit T1 fonts
\usepackage{hyperref}       % hyperlinks
\usepackage{url}            % simple URL typesetting
\usepackage{booktabs}       % professional-quality tables
\usepackage{amsfonts}       % blackboard math symbols
\usepackage{nicefrac}       % compact symbols for 1/2, etc.
\usepackage{microtype}      % microtypography

\usepackage{xspace}
\usepackage{algorithm}
\usepackage{amsmath}
\usepackage{amsthm}
\usepackage{graphicx}
\usepackage{subcaption}
\usepackage{dblfloatfix}
\usepackage{multirow}
\usepackage{xcolor}
\usepackage{multicol}
\usepackage{wrapfig}
\usepackage{pgfplots}
\usepackage{listings}
\pgfplotsset{compat=1.8}
\usepackage{pgfplotstable}
\usepgfplotslibrary{groupplots}
\usepackage{tikz}
\usetikzlibrary{patterns}

\usepackage[noend]{algorithmic}
\newtheorem{theorem}{Theorem}

\newtheorem{loopinvariant}{Loop Invariant}

\newcommand{\printfnsymbol}[1]{%
  \textsuperscript{\@fnsymbol{#1}}%
}

\newcommand{\arxivonly}[1]{}

\newcommand\unknown{\ensuremath{\mathsf{unknown}}\xspace}
\newcommand\robust{\ensuremath{\mathsf{robust}}\xspace}
\newcommand\notrobust{\ensuremath{\mathsf{not\_robust}}\xspace}

\newcommand\unk{{\ensuremath{\mathsf{U}}\xspace}}
\newcommand\rob{{\ensuremath{\mathsf{R}}\xspace}}
\newcommand\norob{{\ensuremath{\mathsf{NR}}\xspace}}
\newcommand\timeout{{\ensuremath{\mathsf{TO}}\xspace}}

\newif\ifsubmit
\ifsubmit
  \newcommand{\todo}[1]{}
\else
  \newcommand{\todo}[1]{\textcolor{red}{TODO: #1}}
\fi

\newcommand{\norm}[1]{\|#1\|}

\newcommand{\ap}{\ensuremath{A}\xspace}
\newcommand{\ar}{\ensuremath{\bar A}\xspace}

\newcommand{\modelfn}{f}
\newcommand{\modelclass}{F\xspace}

\title{Fast Geometric Projections for \\Local Robustness Certification}

\author{%
  Aymeric Fromherz\thanks{First two authors have equal contributions} \\
  Carnegie Mellon University\\
  Pittsburgh, PA, USA \\
  \texttt{fromherz@cmu.edu} \\
  \And
  Klas Leino$^*$ \\
  Carnegie Mellon University\\
  Pittsburgh, PA, USA \\
  \texttt{kleino@cs.cmu.edu} \\
  \AND
  Matt Fredrikson \\
  Carnegie Mellon University\\
  Pittsburgh, PA, USA \\
  \texttt{mfredrik@cmu.edu} \\
  \And
  Bryan Parno \\
  Carnegie Mellon University\\
  Pittsburgh, PA, USA \\
  \texttt{parno@cmu.edu} \\
  \And
  Corina P\u{a}s\u{a}reanu \\
  Carnegie Mellon University\\
  and NASA Ames\\
  Moffett Field, CA, USA \\
  \texttt{pcorina@cmu.edu} \\
}

\iclrfinalcopy % Uncomment for camera-ready version, but NOT for submission.
\begin{document}

\maketitle

\begin{abstract}

% !TEX root=./paper.tex

Local robustness ensures that a model classifies all inputs within an $\ell_p$-ball consistently, which precludes various forms of adversarial inputs.
In this paper, we present a fast procedure for checking local robustness in feed-forward neural networks with piecewise-linear activation functions.
Such networks partition the input space into a set of convex polyhedral regions in which the network’s behavior is linear; 
hence, a systematic search for decision boundaries within the regions around a given input is sufficient for assessing robustness.
% Many existing techniques perform this search using expensive constraint solving.
Crucially, we show how the regions around a point can be analyzed using simple geometric projections, thus admitting an efficient, highly-parallel GPU implementation that excels particularly for the $\ell_2$ norm, where previous work has been less effective.
Empirically we find this approach to be far more precise than many approximate verification approaches, while at the same time performing multiple orders of magnitude faster than complete verifiers, and scaling to much deeper networks.
An implementation of our proposed algorithm is available on GitHub\footnote{Code available at \url{https://github.com/klasleino/fast-geometric-projections}}.

\end{abstract}

\section{Introduction}

% !TEX root=./paper.tex

% {\color{red}
% \begin{itemize}
% \item
% 	Background on adversarial examples, robustness with usual citations.
% \item
% 	Define local robustness.
% \end{itemize}
% }%
We consider the problem of verifying the \emph{local robustness} of piecewise-linear neural networks for a given $\ell_p$ bound.
Precisely, given a point, $x$, network, $F$, and norm bound, $\epsilon$, this entails determining whether Equation~\ref{eq:local-robustness} holds.
\begin{equation}
\label{eq:local-robustness}
\forall x' . \|x-x'\|_p \le \epsilon \Longrightarrow F(x) = F(x')
\end{equation}
This problem carries practical significance, as such networks have been extensively shown to be vulnerable to \emph{adversarial examples}~\citep{PapernotMJFCS16,SzegedyZSBEGF13}, wherein small-norm perturbations are chosen to cause arbitrary misclassifications.
Numerous solutions have been proposed to address variants of this problem.
These can be roughly categorized into three groups: learning rules that aim for robustness on known training data~\citep{croce19mmr,madry2018towards,wong2017provable,zhang19trades,madry19relu}, post-processing methods that provide stochastic guarantees at inference time~\citep{cohen19smoothing,lecuyer2018certified}, and network verification~\citep{balunovic2019geometric,Cheng17Resilience,Dutta2018OutputRA,Ehlers_2017,Fischetti18MILP,gowal18ibp,ProvableBallFitting,katz17reluplex,katz19marabou,singh2018robustness,tjeng18MIP,neurify,weng18fastlip}.
% While approximate verification methods have shown promise in scaling to larger networks, they may introduce an additional penalty to robust accuracy by flagging non-adversarial points, thus limiting their application in practice.
% Exact methods impose no such penalty, but as they rely on expensive constraint-solving techniques, they often do not scale to even moderately-sized networks.

We focus on the problem of network verification---for a given model and input, determining if Equation~\ref{eq:local-robustness} holds---particularly for the $\ell_2$ norm.
Historically, the literature has primarily concentrated on the $\ell_\infty$ norm, with relatively little work on the $\ell_2$ norm;
%, as constraint-solving (commonly used in verification tools) in Euclidean space requires a non-linear objective function, and cannot make as effective use of interval bound propagation.
indeed, many of the best-scaling verification tools do not even support verification with respect to the $\ell_2$ norm.
Nonetheless, the $\ell_2$ norm remains important to consider for ``imperceptible'' adversarial examples~\citep{rony2019l2}.
Furthermore, compared to the $\ell_\infty$ norm, efficient verification for the $\ell_2$ norm presents a particular challenge, as constraint-solving (commonly used in verification tools) in Euclidean space requires a non-linear objective function, and cannot make as effective use of interval-bound propagation.

Existing work on verifying local robustness for the $\ell_2$ norm falls into two primary categories:
\emph{(1)}~expensive, but exact decision procedures, e.g., GeoCert~\citep{ProvableBallFitting} and MIP~\citep{tjeng18MIP}, or
\emph{(2)}~fast, but approximate techniques, e.g., FastLin/CROWN~\citep{weng18fastlip,zhang18crown}.
While approximate verification methods have shown promise in scaling to larger networks, they may introduce an additional penalty to robust accuracy by flagging non-adversarial points, thus limiting their application in practice.
Exact methods impose no such penalty, but as they rely on expensive constraint-solving techniques, they often do not scale well to even networks with a few hundred neurons.

In this paper, we focus on bridging the gap between these two approaches.
In particular, we present a verification technique for Equation~\ref{eq:local-robustness} that neither relies on expensive constraint solving nor conservative over-approximation of the decision boundaries.
\arxivonly{We build on the observation that in Euclidean space, the angle between two random vectors of equal norm is with high probability $\pi/2 \pm O(1/\sqrt{m})$~{\citep[Chapter\ 2]{Blum2020}}.
In other words, random vectors in high dimension are likely to be almost orthogonal, so the distance from a point to a hyperplane segment can often be approximated well by the distance from the point to its projection onto the corresponding hyperplane.}
% \todo{\textbf{Klas:} should we explain this further/include a figure?}

% {\color{red}
% \begin{itemize}
% \item
% 	List contributions: fast projection-based algorithm (plus proof), modification for getting lower bound, faster than previous comparable approaches, etc.
% \end{itemize}
% }%

% Our algorithm (Section~\ref{sec:algorithm}) leverages this insight to exhaustively search the model's decision boundaries around a point \emph{using only geometric projections}.
Our algorithm (Section~\ref{sec:algorithm}) leverages simple projections, rather than constraint solving, to exhaustively search the model's decision boundaries around a point.
The performance benefits of this approach are substantial, especially in the case of $\ell_2$ robustness, where constraint solving is particularly expensive while Euclidean projections can be efficiently computed using the dot product and accelerated on GPU hardware. However, our approach is also applicable to other norms, including $\ell_\infty$ (Section~\ref{sec:eval:linf}).
Our algorithm is embarassingly parallel, and straightforward to implement with facilities for batching that are available in many popular ML libraries.
Additionally, we show how the algorithm can be easily modified to find certified lower bounds for $\epsilon$, rather than verifying a given fixed value (Section~\ref{sect:lower-bound}).

Because our algorithm relies exclusively on projections, it may encounter scenarios in which there is evidence to suggest non-robust behavior, but the network's exact boundaries cannot be conclusively determined without accounting for global constraints (Section~\ref{sec:algorithm}, Figure~\ref{fig:inconclusive}).
In such cases, the algorithm will return \unknown (though it would be possible to fall back on constraint solving).
However, we prove that if the algorithm terminates with a \robust decision, then the model satisfies Equation~\ref{eq:local-robustness}, and likewise if it returns \notrobust, then an adversarial example exists (Theorem~\ref{thm:correctness}).
Note that unlike prior work on approximate verification, our approach can often separate \notrobust cases from \unknown, providing a concrete adversarial example in the former.
In this sense, the algorithm can be characterized as \emph{sound} but \emph{incomplete}, though our experiments show that in practice the algorithm typically comes to a decision.

We show that our implementation outperforms existing exact techniques~\citep{ProvableBallFitting,tjeng18MIP} by multiple orders of magnitude (Section~\ref{sec:evallocal}, Table~\ref{table:robustness} and Section~\ref{sec:eval:linf}), while rarely being inconclusive on instances for which other techniques do not time out.
Moreover, we find our approach enables \emph{scaling to far deeper models} than prior work --- a key step towards verification of networks that are used in practice.
Additionally, on models that have been regularized for efficient verification~\citep{croce19mmr,madry19relu}, our technique performs even faster, and scales to much larger models --- including convolutional networks --- than could be verified using similar techniques (Section~\ref{sec:evallocal}, Table~\ref{table:robustness}).
Finally, we compare our work to approximate verification methods (Section~\ref{sec:evalbound}).
We find that while our implementation is not as fast as previous work on efficient lower-bound computation for large models~\citep{weng18fastlip}, our certified lower bounds are consistently tighter, and in some cases minimal (Section~\ref{sec:evalbound}, Table~\ref{table:lowerbound}).

\section{Algorithm}

% !TEX root=./paper.tex

%(In this section, we only present a high-level view of the algorithm, assuming that we can "collect constraints" and so on. How to do it will be explained in the implementation section)
\label{sec:algorithm}
% \todo{AF: If we remove the termination section, also remove the two mentions of termination here}
In this section we give a high-level overview of our proposed algorithm\arxivonly{ and argue for its correctness (Section~\ref{sec:algorithm:thealg})}, and present some implementation heuristics that improve its performance (Section~\ref{sec:algorithm:heuristics}).
%We begin with a naive version, which captures the basic exhaustive search (Section~\ref{sec:algorithm:thealg}), and then demonstrate how to leverage the compositional structure of the network to optimize the analysis (Section~\ref{sec:alg:tree}).
We also propose a variant (Section~\ref{sect:lower-bound}) to compute certified lower bounds of the robustness radius. 
Correctness proofs for all of the algorithms discussed in this section are provided in Appendix~\ref{sec:proofs}.
% We also discuss additional implementation heuristics in Appendix~\ref{sec:heuristics}.
Because our algorithm applies to arbitrary $\ell_p$ norms, we use the un-subscripted notation $\|\cdot\|$ to refer to a general $\ell_p$ norm for the remainder of this section.
% However, in our evaluation (Section~\ref{sect:eval}), we focus on local robustness for the $\ell_2$ norm.
% We do this for the following two reasons:
% \emph{(1)} the algorithm relies on projections that can be computed efficiently in Euclidean spaces, and
% \emph{(2)} there are few existing tools that can efficiently guarantee local robustness with respect to $\ell_2$, as they are usually limited to processing linear computations (see Section~\ref{sect:related}).

%that the $\ell_2$ norm is induced by an inner product and the ReLU activations are continuous and piecewise linear.
\subsection{The Basic Fast Geometric Projections Algorithm}\label{sec:algorithm:thealg}

% \todo{AF: I think the algorithm itself is norm-independent. The part about $\ell_2$ should go in the implementation section. CP: tried to address pls check}
We propose the \emph{Fast Geometric Projections} (FGP) algorithm, which takes a model, $\modelclass$, an input, $x$, and a bound, $\epsilon$, and either proves that Equation~\ref{eq:local-robustness} is satisfied (i.e., $\epsilon$-local robustness), finds an adversarial input at distance less than $\epsilon$ from $x$, or returns \unknown.
Our algorithm assumes $\modelclass(x) = \text{argmax}\{\modelfn(x)\}$, where $\modelfn : \mathbb{R}^m \to \mathbb{R}^n$ is the function computed by a neural network composed of linear transformations with ReLU activations (i.e., $\modelfn$ is a feed-forward ReLU network).

The algorithm relies on an analysis of all possible \emph{activation regions} around $x$. 
An activation region is a maximal set of inputs having the same \emph{activation pattern} of ReLU nonlinearities.
% , as formalized in Definition~\ref{def:act-pattern-region}.

% \begin{definition}[Activation pattern, region, constraint]
% \label{def:act-pattern-region}
Formally, let $\modelfn_u(x)$ denote the pre-activation value of neuron $u$ in network $\modelclass$ when evaluating $x$.
We say that neuron $u$ is \emph{activated} if $\modelfn_u(x) \geq 0$.
An \emph{activation pattern}, \ap, is a Boolean function over neurons that characterizes whether each neuron is activated.
% $\ap: \mathit{neuron} \rightarrow \mathit{bool}$, is a function that, for each neuron in the network, 
% returns $\mathit{true}$ if and only if $\modelfn_n(x) \geq 0$ (i.e., the neuron is \emph{activated}).
% , and $\mathit{false}$ if it is not. 
Then the \emph{activation region}, \ar, associated with pattern \ap is the set of inputs that realize \ap:
$
\ar := \{ x \mid \forall u.(\modelfn_u(x) \geq 0) \iff \ap(u) \}
$.

Because we assume that $\modelclass$ is a piecewise-linear composition of linear transformations and ReLU activations, we can associate the activation status $\ap(u)$ of any neuron with a closed half-space of $\mathbb{R}^m$~\citep{ProvableBallFitting}.
The \emph{activation constraint}, $C_u$, for neuron $u$ and pattern \ap is the linear inequality $w_u^Tx + b_u \le 0$, where $w_u$ and $b_u$ satisfy
$
\forall x\in\mathbb{R}^m.w_u^Tx + b \le 0 \iff \ap(u)
$.
The coefficients, $w_u$ are equal to the gradient of $\modelfn_u$ with respect to its inputs, evaluated at a point in $\ar$. 
Crucially, because the gradient is the same at every point in $\ar$, the constraints can be computed \emph{from the activation pattern alone} via backpropagation.
More details on this computation are given in Appendix~\ref{sec:appendix:constraints}.

The intersection of these constraints yields the activation region $\ar$, and the facets of \ar correspond to the non-redundant constraints.
The convexity of activation regions follows from this observation, as does the fact that the \emph{decision boundaries} are also linear constraints of the form $\modelfn(x)_i \geq \modelfn(x)_j$ for classes $i$ and $j$.

\begin{figure}[t]
\centering
\begin{subfigure}[t]{0.59\textwidth}
  \centering
  \includegraphics[page=1, width=0.33\textwidth]{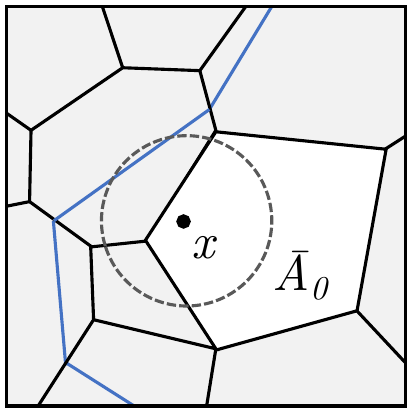}%
  \includegraphics[page=2, width=0.33\textwidth]{figures/fgp_example.pdf}%
  \includegraphics[page=3, width=0.33\textwidth]{figures/fgp_example.pdf}%
  \caption{}
  \label{fig:fgp_example}
\end{subfigure}
\begin{subfigure}[t]{0.25\textwidth}
\end{subfigure}
\begin{subfigure}[t]{0.39\textwidth}
  \centering
  \includegraphics[page=2, width=0.5\textwidth]{figures/inconclusive.pdf}%
  \includegraphics[page=3, width=0.5\textwidth]{figures/inconclusive.pdf}%
  \caption{}
  \label{fig:inconclusive}
\end{subfigure}%
\caption{
  % \textbf{(\subref{fig:inconclusive})} Illustration of the three cases FGP can encounter when discovering a boundary constraint, $C_B$, within the $\epsilon$-ball about $x$.
  % In the case on the \emph{left}, local robustness is not satisfied in $\ar$, and $p\in\ar$ is an adversarial example, so we can return \notrobust. 
  % In the \emph{center} case, local robustness is not satisfied in $\ar$, but $p\notin\ar$ so we cannot distinguish this case from the case on the \emph{right}, in which local robustness \emph{is} satisfied in $\ar$ (and therefore $p$ may not be an adversarial example). 
  % Thus, in both the \emph{center} and \emph{right} cases we must return \unknown.
  \textbf{(\subref{fig:fgp_example})} Illustration of the FGP algorithm.
  We begin in $\ar_0$ \emph{(left)}. 
  We see that activation constraints, $C_1$ and $C_2$ are in the $\epsilon$-ball, thus we enqueue $\ar_1$ and $\ar_2$ \emph{(center)}.
  When searching $\ar_1$, we see that a decision boundary, $C_B$, is within the $\epsilon$-ball. The projection $p$ onto $C_B$ is an adversarial example; we return \notrobust \emph{(right)}.
  \textbf{(\subref{fig:inconclusive})} Illustration of the cases requiring FGP to return \unknown when analyzing a boundary constraint, $C_B$, found within the $\epsilon$-ball about $x$. The true boundary is shown in solid blue; the infinite prolongation of the boundary we are projecting against is shown in dotted blue. 
  In each case the projection, $p$, lies outside of $\ar$, and is in fact not an adversarial example.
  In the case on the \emph{left}, an adversarial example, $a$, exists in $\ar$, while on the \emph{right} local robustness is satisfied in $\ar$.
  However, we cannot distinguish between these cases as $p\notin\ar$ in both; in both cases we must return \unknown.
  % \textbf{(\subref{fig:fgp_bad_cases})} Illustration of two cases where FGP searches unnecessary regions. 
  % \emph{(left)} $C$ is the boundary between regions $\bar{A}_1$ and $\bar{A}_2$. Though $\bar{A}_2$ is not in the $\epsilon$-ball, it will be added to the queue when searching $\bar{A}_1$, since $C$ intersects the $\epsilon$-ball.
  % \emph{(right)} view of four activation constraints viewed from region $\bar{A}_1$. $C$ is implied by the other three constraints, thus the region obtained by flipping the neuron corresponding to $C$ is infeasible, but this region will be searched because $C$ intersects the $\epsilon$-ball.
  % \textbf{(\subref{fig:tree})} Illustration of the "tree" of activation pattern prefixes for the network shown on the left.
  % Each leaf corresponds to an activation pattern, each level corresponds to a layer, and internal nodes correspond to activation pattern prefixes at the respective layer.
}
\end{figure}

The FGP algorithm performs a search of all activation regions that might be at distance less than $\epsilon$ from $x$.
We begin by analyzing the region, $\ar_0$, associated with the activation pattern of the input, $x$, as follows.

First, we check to see if $\ar_0$ contains a decision boundary, $C$, that is at distance less than $\epsilon$ from $x$; if so, we take the projection, $p$, from $x$ onto $C$.
Intuitively, the projection is an input satisfying $C$ that is at minimal distance from $x$, i.e., $\forall x'.C(x') \Longrightarrow \| x-p\| \leq \|x-x'\|$.
Similarly we define the distance, $d$, from $x$ to a constraint or decision boundary, $C$, as $d(x, C) := \min_{x' : C(x')}{\norm{x - x'}}$, i.e., $d$ is the distance from $x$ to its projection onto $C$.
If $p$ does not have the same class as $x$ according to $\modelclass$ (or lies directly on the decision boundary between classes), then $p$ is an adversarial example.
If $p$ has the same class as $x$, it means that the projection to the decision boundary is outside of the activation region we are currently analyzing;
however, this is not sufficient to conclude that \emph{no} point on the decision boundary is inside both the current activation region and the $\epsilon$-ball (see Figure~\ref{fig:inconclusive}). 
Therefore, we return \unknown.

Otherwise, we collect all activation constraints in $\ar_0$;
there is one such activation constraint per neuron $u$ in the network; thus each constraint corresponds to the neighboring region, $\ar^u_0$, which has the same activation pattern as $\ar_0$, except with neuron $u$ flipped, i.e., $\ap_0(u') \neq \ap_0^u(u') \Longleftrightarrow u = u'$.
For each constraint $C_u$, we check whether it might be at distance less than $\epsilon$ from $x$ using a geometric projection.
Although $C_u$ is only valid in the polytope corresponding to the activation region $\ar_0$, we compute the distance to the hyperplane corresponding to the infinite prolongation of $C_u$.
As such, we only obtain a lower bound on the distance from $x$ to $C_u$.
If the computed distance is smaller than $\epsilon$, we enqueue the corresponding activation region if it has not been searched yet.
% Here we define the distance, $d$, from $x$ to a constraint or decision boundary, $C$, as $d(x, C) := \min_{x' : C(x')}{\norm{x - x'}}$, i.e., $d$ is the distance from $x$ to its projection onto $C$.
Thus the queue contains unexplored regions that might be at distance less than $\epsilon$ from $x$. 
The algorithm continues to analyze each enqueued region in the same way, each explored region enqueuing neighbouring regions that might be at distance less than $\epsilon$ from $x$,
until the queue is empty.
Exhausting the queue means that we did not find any adversarial examples in any activation region that might intersect with the $\epsilon$-ball centered at $x$. 
We therefore conclude that $x$ is locally $\epsilon$-robust.
An illustration of a simple execution of this algorithm is shown in Figure~\ref{fig:fgp_example}.
%
% Note that since there are finitely many activation regions and we only explore each region at most once, this algorithm terminates.

% \begin{wrapfigure}{r}{0.47\textwidth}
% \raisebox{0pt}[\dimexpr\height-1.0\baselineskip\relax]{
% \begin{minipage}{0.47\textwidth}
% \begin{algorithm}[H]
%   \caption{~~\small Certified Lower Bounds}
%   \label{alg:lowerbounds}
% \begin{algorithmic}
%   \scriptsize
%   \STATE \textbf{Input:} model $\modelclass$, data $x$, bound $\epsilon$
%   \STATE Initialize $queue = priorityQ()$, $visited = \emptyset$, $\beta = 0$
%   \STATE \ldots
%   \WHILE{$queue \neq \emptyset$}
%     \STATE $C$ = \deq($queue$)
%     \STATE $\beta$ = max($\beta$, $d(x, C)$)
%     \IF{\isdb($C$)}
%       \STATE \textbf{return} $\beta$
%     \ELSE
%       \STATE \ldots 
%     \ENDIF
%   \ENDWHILE
%   \STATE \textbf{return} $\epsilon$
% \end{algorithmic}
% \end{algorithm} 
% \end{minipage}%
% }%
% \vspace{-1em}
% \end{wrapfigure}

\paragraph{Conditions on Adversarial Examples.}

It is common in practice to include domain-specific conditions specifying which inputs are considered valid adversarial examples.
For example, in the image domain, adversarial examples are typically subject to \emph{box constraints} requiring each pixel to be in the range $[0,1]$.
These constraints can easily be incorporated into the FGP algorithm by checking them when we check if $F(x)\neq F(p)$, when a decision boundary is found.
If $p$ is not a valid adversarial example, the algorithm would return \unknown.

\paragraph{Correctness.}

We show that when the FGP algorithm returns \notrobust, there exists an adversarial example, and when it returns \robust, the model is locally robust at $x$, with radius $\epsilon$. However, the algorithm may also return \unknown, in which case we do not claim anything about the robustness of the model. A complete proof of Theorem~\ref{thm:correctness} is provided in Appendix~\ref{sec:fgp_proof}.

\begin{theorem}\label{thm:correctness}
(1) When the FGP algorithm returns \notrobust, there exists an adversarial example, $p$, such that $\norm{x - p} \leq \epsilon$ and $\modelclass(x) \neq \modelclass(p)$. 
(2) When it returns \robust, Equation~\ref{eq:local-robustness} holds.
\end{theorem}

\subsection{Heuristics for FGP}
\label{sec:algorithm:heuristics}

To improve the performance and precision of our algorithm, we use several additional heuristics.
%In this section we summarize these heuristics and the observations that give rise to them.
%In some cases, the heuristics present a trade-off, and thus we provide the option to use or not use them in our implementation.

% \noindent{\bf Analyzing Decision Boundaries First.}
% The correctness of the FGP algorithm is independent of how we iterate over the queue of boundaries and regions to analyze.
% As such, we can prioritize the most relevant constraints, i.e. decision boundaries.
% Note, however, that this optimization cannot be used when searching for a certified lower bound as this variant of the algorithm requires the constraints to be searched in order of distance to $x$.

%\paragraph{
\noindent{\bf Batching.}
%Most deep learning software is designed to easily enable efficient batch computations on a GPU.
%In particular, 
The gradient backpropagations for computing constraints and the dot product projections for calculating distances lend themselves well to batching on the GPU.
%We leverage this 
In our implementation we dequeue multiple elements from the queue at once and calculate the constraints and constraint distances for the corresponding regions in parallel, leading to
%We find that this optimization enables 
a speedup of up to 2x on large examples for a well-chosen batch size ($\sim$10-100).

%\paragraph{
\noindent{\bf Caching Intermediate Computations.}
%As explained previously, an activation constraint for a neuron $u$ at layer $N$ in a region $\bar{A}$ is entirely determined by the %activation
%pattern prefix of all neurons up to layer $N$ excluded in $\bar{A}$.
%Thus, for any two regions $\bar{A}, \bar{A'}$ with the same activations up to layer $N'$, constraints for neurons in layers up to $N'$ %are identical
%in $\bar{A}$ and $\bar{A'}$. As such, they do not need to be recomputed when exploring $\bar{A}$ and $\bar{A'}$ successively.
%Always caching such constraints is a delicate matter. A first approach would be to always explore fully the subtree of an activation %pattern prefix before switching to another subtree,
%which would avoid recomputation of constraints corresponding to layers in the prefix. Unfortunately, this might prevent efficient %parallelism,
%the extreme case being if only one new region is discovered in the subtree at each iteration.
%On the other end of the spectrum, one could preserve the existing strong parallelism, and cache for each activation pattern prefix %encountered
%all constraints that do not need to be recomputed. But this would require a large amount of memory, making the analysis intractable for %larger networks.
%Finding the optimal tradeoff between parallelism, memory usage, and efficiency is tricky, and likely depends on the analyzed network %and input point.
%In this work, we instead propose a generic approach that always improves on a naive implementation. 
We observe that activation constraints for the
first layer are invariant across all regions, as there is no earlier neuron to depend on. 
Thus, we can precompute all such constraints once for all inputs to be analyzed.

%\paragraph{
\noindent{\bf Exploring the Full Queue.}
If the analysis of a decision boundary is inconclusive, the FGP algorithm, as presented, stops and returns \unknown.
We also implement a variant where we instead record the presence of an ambiguous decision boundary but continue exploring other regions in the queue.
If we empty the queue after unsuccessfully analyzing a decision boundary, we nevertheless return \unknown.
However, oftentimes a true adversarial example is found during the continued search, allowing us to return \notrobust conclusively.
%The correctness of our algorithm still holds in this variant:
%an inconclusive analysis of a decision boundary leads to either returning \unknown, or exhibiting a true adversarial example found %subsequently.

%{\bf corina: I would cut this paragraph since we do not use it in eval?}
%%\paragraph{
%\noindent{\bf Falling Back on LP Solving.}
%Complete tools \cite{katz17reluplex,katz19marabou,ProvableBallFitting,tjeng18MIP,Fischetti18MILP,Cheng17Resilience} often rely heavily on solvers to analyze models. While in our work we aim to minimize the use of constraint solving, which is computationally expensive,
%we can use a solver sparingly, when we cannot rule out adversarial examples in a specific activation region. We use an LP solver, since both the activation constraints and the decision boundaries for a  region are linear constraints.
%Thus, although we can discharge many verification conditions through fast geometric projections, we provide a complete, but slower, variant of our algorithm through the use of LP solving. We do not use this option in our evaluation. 
%

\begin{table}[t]
\centering
% !TEX root=../paper.tex

\begin{tabular}{l|cc|cc}
\toprule
\multicolumn{1}{c}{} & 
\multicolumn{2}{c}{Vanilla} & 
\multicolumn{2}{c}{Full Queue} \\

\multicolumn{1}{c}{Model} & 
\multicolumn{1}{c}{Time (s)} & 
\multicolumn{1}{c}{\unk} & 
\multicolumn{1}{c}{Time (s)} &
\multicolumn{1}{c}{\unk} \\

\midrule
mnist20x3  & 0.012 & 7  & 0.012 & 4 \\ 
mnist20x6  & 0.177 & 8  & 0.174 & 5 \\
mnist20x9  & 4.897 & 16 & 7.389 & 3 \\
mnist40x3  & 5.417 & 9 & 6.712 & 3 \\ 
% fmnist20x3 & 0.035 & 7  & 0.036 & 2 \\ 
% fmnist40x3 & 1.528 & 9  & 2.910 & 1 \\ 
\bottomrule
\end{tabular}%

\caption{Comparison of FGP and its variant with full exploration of the queue}
\label{table:keepgoing}
\end{table}

Table~\ref{table:keepgoing} displays the results of experiments evaluating the impact of exploring the entire search queue instead of stopping at the first decision boundary (as described in Section~\ref{sec:algorithm:thealg}).
The experimental results show that this heuristic decreases the number of \unknown results by approximately $50\%$ while only having a minor impact on the speed of the execution.
%\todo{check this:}
Moreover, we reduce the number of \unknown results to the extent that we recover the results obtained by GeoCert in every instance for which GeoCert terminates, except 3 non-robust points on mnist40x3, while nevertheless performing our analysis two to three orders of magnitude more quickly.

\subsection{Certified Lower Bounds}
\label{sect:lower-bound}

%\todo{AF: Move this at the end of section 2, and provide a snippet of the modified algorithm + give an intuition regarding why this is correct}
%\todo{Bryan: Agreed, but adjust the description of the queue, since at that point, you haven't said that it's two queues internally}
We now consider the related problem of finding a \emph{lower bound} on the local robustness of the network.
A variant of the FGP algorithm can provide certified lower bounds by using a priority queue;
constraints are enqueued with priority corresponding to their distance from $x$, such that the closest constraint is always at the front of the queue.
We keep track of the current certified lower bound in a variable, $\beta$. 
At each iteration, we set $\beta$ to the maximum of the old $\beta$ and the distance from $x$ to the constraint at the front of the queue.

The algorithm terminates either when all constraints at distance less than the  initially specified $\epsilon$ were handled,
or when a decision boundary for which the initial algorithm returns \unknown or \notrobust is found. 
In the first case, we return $\epsilon$; in the second, we return the value stored in $\beta$ at this iteration.

The proof of this variant is similar to the proof of the FGP algorithm.
It relies on the following loop invariant, which we prove in Appendix~\ref{sec:loop_proof}.

\begin{loopinvariant}\label{thm:loopinvariant}
% At each iteration 
(1) All activation regions at distance less than $\beta$ from $x$ were previously visited, (2) $\beta$ is always smaller than $\epsilon$,
and (3) there is no adversarial point at distance less than $\beta$.
\end{loopinvariant}

% \section{Efficient Implementation}
% \label{sect:implementation}

% \input{implementation}

% \input{heuristics}

\section{Evaluation}
\label{sect:eval}

% !TEX root=./paper.tex

In this section, we evaluate the performance of our implementation of the FGP algorithm and its variant for computing lower bounds.
Section~\ref{sec:evallocal} discusses the performance of FGP compared to existing tools that perform exact $\ell_2$ robustness verification.
Section~\ref{sec:evalbound} compares the lower bounds certified by our FGP variant from Section~\ref{sect:lower-bound} to those of other approximate certification methods.
In short, we find that our approach outperforms existing exact verification tools by several orders of magnitude (2-4), and produces more precise lower bounds (3-25 times larger) than the relevant work on approximate certification.

Additionally, in Section~\ref{sec:evallocal}, we explore the scalability of our algorithm. 
%--- particularly with respect to the depth of the network being verified, finding that pruning based on the network's compositional structure (Section~\ref{sec:alg:tree}) is particularly useful for scaling to deeper networks.
We find that when the networks are strongly regularized for verifiability, our approach even scales to CNNs.
Finally, we remark on the flexibility of FGP with respect to the norm in Section~\ref{sec:eval:linf}, and observe that our approach is also faster than existing complete verification tools
when performing $\ell_\infty$ robustness verification.

We performed experiments on three standard datasets: MNIST, Fashion-MNIST, and CIFAR10. 
We evaluated both on models trained for robustness using adversarial training~\citep{madry2018towards}, and on models trained for verifiability \emph{and} robustness using maximum margin regularization (MMR)~\citep{croce19mmr} or ReLU Stability (RS)~\citep{madry19relu}.
We refer to each model by the dataset it was trained on, followed by the architecture of the hidden layers.
For example, ``mnist20x3'' refers to a model trained on MNIST with 3 hidden layers of 20 neurons each.
The ``cnn'' architecture refers to a common CNN architecture used to benchmark CNN verification in the literature; details are given in Appendix~\ref{sec:appendix:hyper}.
Models marked with ``*'' were trained with MMR; models marked with ``\dag'' were trained with RS; all other models were trained using PGD adversarial training~\citep{madry2018towards}.
The hyperparameters used for training are given in Appendix~\ref{sec:appendix:hyper}.

In each experiment, measurements are obtained by evaluating on 100 arbitrary instances from the test set.
All experiments were run on a 4.2GHz Intel Core i7-7700K with 32 GB of RAM, and a Tesla K80 GPU with 12 GB of RAM.
% Our implementation is written in Python using the Keras framework~\citep{keras}.
% with the TensorFlow backend~\cite{tensorflow}. 

\subsection{Local Robustness Certification}\label{sec:evallocal}

\begin{table*}[t]
\centering
\begin{subfigure}{1.0\textwidth}
\centering
\resizebox{\textwidth}{!}{
% !TEX root=../paper.tex

\begin{tabular}{l|cccccc|ccccc|ccccc|c}
\toprule
\multicolumn{1}{c}{} &
% \multicolumn{3}{c}{TreeFGP} &
\multicolumn{6}{c}{FGP} & 
\multicolumn{5}{c}{GeoCert} & 
\multicolumn{5}{c}{MIP} & 
\multicolumn{1}{c}{} \\

Model & 
Time (s) & \rob & \norob & \unk & \timeout & VRA &
Time (s) & \rob & \norob & \timeout & VRA &
Time (s) & \rob & \norob & \timeout & VRA & 
$\text{VRA}_\text{UB}$ \\ \midrule

mnist20x3 &
	0.012 & 87 & 8 & 4 & 1 & 0.82 & 
	% 0.022} & 87 & 82 &
	10.99 & 82 & 9 & 9 & 0.77 &
	97.02 & 58 & 9 & 33 & 0.54 &
	0.84 \\
mnist20x6 & 
	0.174 & 79 & 9 & 5 & 7 & 0.74 &
	% 0.332} & 77 & 72 &
	78.93 & 49 & 9 & 42 & 0.44 &
	>120 & 0 & 0 & 100 & 0.00&
	0.85 \\
mnist20x9 & 
	7.389 & 44 & 16 & 3 & 37 & 0.42 &
	% 7.952} & 42 & 40 &
	>120 & 10 & 10 & 80 & 0.10 &
	>120 & 0 & 0 & 100 & 0.00 &
	0.82 \\
mnist40x3 & 
	6.712 & 54 & 6 & 3 & 37 & 0.51 &
	% 6.619} & 54 & 51 &
	>120 & 16 & 11 & 73 & 0.16 &
	>120 & 0 & 0 & 100 & 0.00 &
	0.89 \\ \midrule \midrule

fmnist200x4* &
	0.025 & 81 & 14 & 4 & 1 & 0.73 &
	% 0.039} & 80 & 71 &
	54.85 & 45 & 14 & 41 & 0.41 &
	>120 & 0 & 0 & 100 & 0.00 &
	0.76 \\
fmnist200x6* &
	0.087 & 66 & 12 & 3 & 19 & 0.60 & 
	% 0.118} & 71 & 64 &
	>120 & 28 & 5 & 67 & 0.26 &
	>120 & 0 & 0 & 100 & 0.00 &
	0.75 \\
fmnist100x20* &
	0.057 & 86 & 7 & 7 & 0 & 0.61 &
	% 0.088} & 71 & 59 &
	>120 & 42 & 8 & 50 & 0.37 &
	>120 & 0 & 0 & 100 & 0.00 &
	0.66 \\ 
\midrule
cifar-cnn$^\dag$ &
	0.058 & 86 & 14 & 0 & 0 & 0.27 &
	\multicolumn{5}{c|}{not supported} & 
	>120 & 0 & 0 & 100 & 0.00 &
	0.27 \\
\bottomrule
\end{tabular}%
}%
\vspace{-.5em}
\caption{}\label{table:robustness}
\end{subfigure}

%\begin{subfigure}{0.47\textwidth}
%\centering
%\scriptsize
%\input{figures/table_keepgoing}
%\vspace{-.5em}
%\caption{}
%\label{table:keepgoing}
%\end{subfigure}%
% 
\begin{subfigure}{1.0\textwidth}
\scriptsize
\centering
%!TEX root=../paper.tex

\begin{tabular}{lccc}
\toprule
 & $\text{FGP}_\text{LB}$ & FastLin & \\ 
\multicolumn{1}{c}{Model} & Mean Bound     & Mean Bound & Median Ratio \\ 
\midrule
fmnist500x4  & 0.124 & 0.078      & 0.329 \\
fmnist500x5  & 0.134 & 0.092      & 0.693 \\
fmnist1000x3 & 0.083 & 0.021      & 0.035 \\ 
\bottomrule
\end{tabular}

\vspace{-.5em}
\caption{}
\label{table:lowerbound}
\end{subfigure}
\vspace{-.5em}
\caption{
	\textbf{(\subref{table:robustness})} Comparison of $\ell_2$ local robustness certification (FGP vs. GeoCert vs. MIP) on 100 arbitrary test instances, including the median runtime, the certification result --- either \robust ({\scriptsize\rob}), \notrobust ({\scriptsize\norob}), \unknown ({\scriptsize\unk}), or a timeout ({\scriptsize\timeout}) --- and the corresponding Verified robust accuracy (VRA).
	The upper bound on the VRA ($\text{VRA}_\text{UB}$) is also given. 
	Each instance is given a time budget of 120 seconds.
	Results are for $\epsilon=0.25$, except on CIFAR10, where we use $\epsilon=0.1$, as was used by ~\citep{croce19mmr}.
%	\textbf{(\subref{table:keepgoing})} Comparison of FGP and its variant with full exploration of the queue (Section~\ref{sec:heuristics}). 
	\textbf{(\subref{table:lowerbound})} Comparison of the lower bound obtained by FGP (Section~\ref{sect:lower-bound}) to that obtained by FastLin.
}
\vspace{-1em}
\end{table*}

We first compare the efficiency of our implementation to that of other tools certifying local robustness.
GeoCert~\citep{ProvableBallFitting} and MIP~\citep{tjeng18MIP} are the tools most comparable to ours as they are able to exactly check for robustness with respect to the $\ell_2$ norm.
We used commit hash 8730aba for GeoCert, and v0.2.1 for MIP.
Specifically, we compare the median run time over the analyses of each of the selected 100 instances, the number of examples on which each tool terminates, the result of the analyses that terminate, and the corresponding verified robust accuracy (VRA), i.e., the fraction of points that were both correctly classified by the model and verified as robust. 
In addition, we report an upper bound on the VRA for each model, obtained by running PGD attacks (hyperparameters included in Appendix~\ref{sec:appendix:hyper}) on the correctly-classified points on which every method either timed out or reported \unknown.
Results are given for an $\ell_2$ norm bound of $\epsilon = 0.25$, with a computation budget of two minutes per instance.
The results for these experiments are presented in Table~\ref{table:robustness}.

We observe that FGP always outperforms GeoCert by two to three orders of magnitude, without sacrificing precision; i.e., we rarely return \unknown when GeoCert terminates. 
MIP frequently times out with a time budget of 120 seconds, indicating that we are faster by at least four orders of magnitude.
This is consistent with the 100 to 1000 seconds per solve on a MNIST model with three hidden layers of 20 neurons each reported by~\citet{tjeng18MIP};
their technique performs best on the $\ell_1$ or $\ell_\infty$ norms.
In addition, we find that FGP consistently verifies the highest fraction of points, and yields the best VRA.
Moreover, on mnist20x3, FGP comes within 2 percentage points of the VRA upper bound.
This suggests that FGP comes close to verifying \emph{every} robust instance on this model. 

\paragraph{Models Trained for Verifiability.}
Adversarial training~\citep{madry2018towards}, used to train the models in Sections~\ref{sec:evallocal} and \ref{sec:evalbound}, attempts to ensure that points on the data manifold will be far from the model's decision boundary.
Even if it achieves this objective, it may nonetheless be difficult to verify because the performance of FGP depends not only on the decision boundaries, but also on the internal activation constraints.
Thus, we expect certification will be efficient only when the points to be certified are also far from internal boundaries, leading to fewer regions to explore.
Recent work has sought to develop training procedures that not only encourage robustness, but that also optimize for efficient verification~\citep{croce19mmr,wong2017provable,madry19relu}.
Maximum margin regularization (MMR)~\citep{croce19mmr} and ReLU stability (RS)~\citep{madry19relu} are particularly suited to our work, as they follow from the same intuition highlighted above.

%\begin{wrapfigure}{r}{0.43\textwidth}
%\vspace{0em}
%\resizebox{0.4\textwidth}{!}{
%\input{figures/regions_graph_depth}}%
%\vspace{0em}
%\caption{
%	Plot of the number of regions explored by FGP and GeoCert, as a function of the depth of the network.
%	The size of the network was held constant at 60 neurons.}
%\label{fig:regions_depth}
%\vspace{-2em}
%\end{wrapfigure}

Using MMR, FGP is able to scale to much larger models with hundreds or thousands of neurons, and tens of layers, as shown in the bottom half of Table~\ref{table:robustness}.
Here again, we see that FGP outperforms the other approaches in terms of both time and successfully-verified points.
By comparison, while GeoCert also experienced improved performance on the MMR-trained models, our method remains over three orders of magnitude faster.
MIP continued to time out in nearly every case.

We found that we were able provide even stronger regularization with RS, allowing us to scale to CNNs (Table~\ref{table:robustness}), which have far more internal neurons than even large dense networks.
We found that these highly regularized CNNs verified more quickly than some of the less-regularized dense networks, though, as with other methods that produce verifiable CNNs, this came at the cost of a penalty on the model's accuracy.

%Cases where our approach returns \unknown correspond to inputs that can be deemed \emph{difficult}, as other tools time o{}ut on those.
%\paragraph{Depth Scalability.}
%FGP searches an overapproximation of the relevant search space for an adversarial example, potentially searching many more regions than are necessary to certify robustness.
%Figure~\ref{fig:regions_depth} plots the number of regions searched by FGP, as compared to those searched by GeoCert, which searches the minimal required set of regions.
%In a single-layer model, FGP searches 10x as many regions as GeoCert, while in the deep model with 10 layers it searches only 3x as many regions, suggesting that FGP becomes \emph{particularly efficient on deeper networks}.
%Note that the cost per region is more than 2,000x less in FGP than in GeoCert.

\subsection{Certified Lower Bounds}\label{sec:evalbound}

We now evaluate the variant of our algorithm computing certified lower bounds on the robustness radius (Section~\ref{sect:lower-bound}). 
To this end, we compare the performance of our approach to FastLin~\citep{weng18fastlip}, which is designed to provide quick, but potentially loose lower bounds on the local robustness of large ReLU networks.
We compare the certified lower bound reported by our implementation of FGP after 60 seconds of computation (on models large enough such that FGP rarely terminates in 60 seconds) to the lower bound reported by FastLin; FastLin always reported results in less than two seconds on the models analyzed. 
The results are presented in Table~\ref{table:lowerbound}.
The mean lower bound is reported for both methods, and we observe that on the models tested, FGP is able to find a better lower bound on average, though it requires considerably more computation time.

Because the optimal bound may vary between instances, we also report the median ratio of the lower bounds obtained by the two methods on each individual instance.
Here we see that FastLin may indeed be quite loose, as on a typical instance it achieves as low as 4\% and only as high as 69\% of the bound obtained by FGP. 
Finally, we note that when FGP terminates by finding a decision boundary, if the projection onto that boundary is a true adversarial example, then \emph{the lower bound is tight}.
In our experiments, there were few such examples --- three on fmnist500x4 and one on fmnist1000x3 --- however, in these cases, the lower bound obtained by FastLin was very loose, achieving 4-15\% of the optimal bound on fmnist500x4, and only 0.8\% of the optimal bound on fmnist1000x3. 
This suggests that while FastLin has been demonstrated to scale to large networks, one must be careful with its application, as there may be cases in which the bound it provides is too conservative.

\subsection{Generalization to Other Norms}\label{sec:eval:linf}

The only aspect of FGP that depends on the norm is the projection and the projected distance computation, making our approach highly modular with respect to the norm.
As such, we added support for $\ell_\infty$ robustness certification by providing a projection and a projected distance computation in FGP.

In this section, we compare the efficiency of this implementation to GeoCert~\citep{ProvableBallFitting} and ERAN~\citep{singh2019krelu}. ERAN uses abstract interpretation to analyze networks using conservative over-approximations. As such, the analysis can certify that an input is robust, but can yield false positives when flagging an input as not robust. In this evaluation, we use the \emph{DeepPoly} abstract domain, and the complete mode of ERAN which falls back on constraint solving when an input is classified as not robust.

Our experimental setup is similar to the one presented in Section~\ref{sec:evallocal}.
For each network, we arbitrarily pick 100 inputs from the test set and report the median analysis time of each tool.
We give each tool a time budget of 120 seconds per input. We report the number of instances
on which each tool terminates within this time frame, as well as the results of the analyses that terminate. Results are given in Table~\ref{table:linf_small} for an $\ell_\infty$ norm bound of $\epsilon = 0.01$, which contains roughly the same volume as the $\ell_2$ ball used in most of our evaluation.

We refer to each model by the dataset it was trained on, followed by the architecture of the hidden layers. For example, ``mnist20x3'' refers to a model trained on MNIST with 3 hidden layers of 20 neurons each.
Models marked with an asterisk were trained with MMR~\citep{croce19mmr}; other models were trained using PGD adversarial training~\citep{madry2018towards}.

\begin{table}[t]
\scriptsize
\centering
% !TEX root=../paper.tex

\begin{tabular}{l|ccccc|cccc|cccc}
\toprule
\multicolumn{1}{c}{} &
% \multicolumn{3}{c}{TreeFGP} &
\multicolumn{5}{c}{FGP} & 
\multicolumn{4}{c}{ERAN+MIP} & 
\multicolumn{4}{c}{GeoCert} \\

Model & 
Time (s) & \rob & \norob & \unk & \timeout &
Time (s) & \rob & \norob & \timeout &
Time (s) & \rob & \norob & \timeout \\ \midrule

mnist20x3 & 
	0.004 & 94 & 0 & 6 & 0 & 
	0.017 & 95 & 5 & 0 &
	0.360 & 95 & 5 & 0  \\
mnist20x6 & 
	0.016 & 95 & 0 & 4 & 1 &
	0.054 & 97 & 3 & 0 &
	0.988 & 97 & 3 & 0 \\
mnist20x9 & 
	0.021 & 84 & 4 & 8 & 4 &
	0.080 & 92 & 8 & 0 &
	2.799 & 89 & 8 & 3 \\
mnist40x3 & 
	0.094 & 91 & 2 & 5 & 2 &
	0.040 & 96 & 4 & 0 &
	2.843 & 94 & 4 & 2 \\
% mnist60x3 & 
% 	 & & & & &
% 	 0.085 & 94 & 6 & 0 &
% 	 21.438 & 72 & 6 & 22 \\ 
\midrule \midrule

fmnist100x6* &
	 0.012 & 85 & 7 & 8 & 0 &
	 0.982 & 91 & 9 & 0 &
	 0.750 & 91 & 9 & 0 \\
fmnist100x10* &
	 0.021 & 92 & 1 & 7 & 0 &
	 2.104 & 94 & 6 & 0 &
	 1.308 & 93 & 6 & 1 \\
%mnist-cnn* &
%	 0.281 & 76 & 0 & 22 & 2 &
%	 3.338 & 100 & 0 & 0 &
%	 \multicolumn{4}{c}{not supported} \\
\bottomrule
\end{tabular}%

\caption{Comparison of $\ell_\infty$ local robustness certification (FGP vs. ERAN vs. GeoCert) on 100 arbitrary test instances with a time budget of 120 seconds, including the median runtime and the certification result: either \robust ({\scriptsize\rob}), \notrobust ({\scriptsize\norob}), \unknown ({\scriptsize\unk}), or a timeout ({\scriptsize\timeout}).
Results are for $\epsilon=0.01$.
}
\label{table:linf_small}
\end{table}

%\begin{subfigure}{1.0\textwidth}
%\centering
%\input{figures/table_linf_large}
%\caption{}
%\label{table:linf_large}
%\end{subfigure}
%

We observe that we consistently outperform GeoCert by about two orders of magnitude, both on models trained with MMR and using adversarial training.
On models trained for robustness (upper rows of Table~\ref{table:linf_small}), FGP is 0.5x to 5x faster than ERAN. In particular, FGP seems to scale better
to deeper networks, while ERAN performs better on wider networks. 
Interestingly, ERAN does not benefit as much from MMR training as FGP and GeoCert. On large models trained with MMR, our tool is one to two orders of magnitude faster
than ERAN, and ERAN is only about 2x faster than GeoCert.

Finally, while FGP is able to determine robustness for almost all inputs deemed robust by GeoCert or ERAN, it struggles to find adversarial examples for non-robust points,
which results in a higher number of \unknown cases compared to the $\ell_2$ variant. This is not highly surprising, as projections are better-suited to the Euclidean space than to $\ell_\infty$ space.
While the projection in Euclidean space is unique, this is not always the case in $\ell_\infty$ space; we pick one arbitrary projection which thus might not lead to an adversarial example.

\section{Related Work}
\label{sect:related}

% !TEX root=./paper.tex

% \todo{Describe related work and present nuanced comparison to methods that are not directly comparable.
% In particular, \cite{katz17reluplex,katz19marabou,weng18fastlip,cohen19smoothing,ProvableBallFitting}.}
Our work can be grouped with approaches for verifying neural networks that aim to check local robustness exactly~\citep{ProvableBallFitting,katz17reluplex,katz19marabou,tjeng18MIP}; the primary difference is that our approach avoids expensive constraint solving at the price of incompleteness.

GeoCert~\citep{ProvableBallFitting} is the closest work to ours; it aims to exactly compute local robustness of deep neural networks for convex norms.
Unlike our approach, GeoCert computes the largest $\ell_p$ ball centered at an input point within which the network is robust.  
Our experimental comparison with GeoCert shows that our approach scales much better. 
This is not surprising as GeoCert relies on projections to polytopes, which are solved by a quadratic program (QP) with linear constraints. This enables to compute exactly the distance from a point to an activation constraint, instead of an underapproximation as our approach does. 
In contrast, our approach uses projections to affine subspaces, which have a simpler, closed-form solution.
\cite{lim2020hierarchical} builds on GeoCert, but reduces the number of activation regions to consider by exploring the network in a hierarchical manner.
Although \citeauthor{lim2020hierarchical} reduce the number of QPs required for verification, their variant still relies on constraint solving.
While they improve on GeoCert by up to a single order of magnitude, our approach consistently outperforms GeoCert by 2-4 orders of magnitude.
MIP~\citep{tjeng18MIP} is an alternative to GeoCert based on mixed-integer programming; it also requires solving QPs for the $\ell_2$ norm. We could fall back on similar techniques to provide a slower, but complete variant of our algorithm when our projections cannot reach a conclusion about a decision boundary.

Reluplex~\citep{katz17reluplex} and its successor, Marabou~\citep{katz19marabou} are complete verification tools based on SMT solving techniques. 
Unlike our approach, Reluplex and Marabou do not support the $\ell_2$ norm. 
% ERAN~\cite{singh2019krelu}, and its predecessor, AI2~\cite{gehr2018ai}, rely on abstract interpretation~\cite{cc-POPL77} to automatically verify local robustness of neural networks against both $\ell_\infty$ and geometric input perturbations.
% As such, they use sound, conservative over-approximations to perform their analysis,
% which leads to false positives, i.e. robust inputs incorrectly classified as not robust.
% In comparison, all of the inputs flagged as \notrobust by our approach have an adversarial example at distance less than $\epsilon$.
AI2~\citep{gehr2018ai} and its successor, ERAN~\citep{singh2018robustness}, are based on abstract interpretation~\citep{cc-POPL77}, using conservative over-approximations to perform their analysis, which leads to false positives, i.e. robust inputs incorrectly classified as not robust.
A mode of ERAN enables complete verification by falling back on a constraint solver when an input is classified as not robust; however
this tool does not support the $\ell_2$ norm either.
%In comparison, all of the inputs flagged as \notrobust by our approach have an adversarial example at distance less than $\epsilon$.
%These methods can be seen as a relaxation of MIP, and are thus likewise better-suited to the $\ell_\infty$ norm.

FastLin~\citep{weng18fastlip} exploits the special structure of ReLU networks to efficiently compute lower bounds on minimal adversarial distortions.
CROWN~\citep{zhang18crown} later expanded this to general activation functions.
Although FastLin has been shown to be very scalable, our experiments indicate that the computed bounds may be imprecise.

Recently, a quite different approach has been proposed for robustness certification. 
Randomized Smoothing~\citep{cohen19smoothing,lecuyer2018certified} is a post-processing technique that provides a stochastic robustness guarantee at inference time.
This approach differs from our approach in that it \emph{(1)} modifies the predictions of the original model (increasing the complexity of making predictions), and \emph{(2)} provides a \emph{probabilistic} robustness guarantee that is quantified via a confidence bound.
% , rather than providing a sound deterministic guarantee.
As such it provides an alternative set of costs and benefits as compared to static verification approaches.
% Its complexity also differs from that of our approach, as it is dependent less on the architecture of the model, but rather on the number of samples required to perform its post-processing of the model's output.
Its complexity also differs from that of FGP, as it is dependent primarily on the number of samples required to perform its post-processing of the model's output.
We find that in our experimental setup, achieving the same probabilistic guarantee as the experiments described in \citep{cohen19smoothing} requires $10^5$ samples, taking approximately 4.5 seconds per instance.
 % to make a certified prediction on each of the models tested in Section~\ref{sec:evallocal}. 
Thus, for the models in our evaluation, FGP is on average faster or comparable in performance.

\section{Conclusion}

% !TEX root=./paper.tex

In this paper, we presented a novel approach for verifying the local robustness of networks with piecewise linear activation functions, that relies neither on constraint solving nor conservative over-approximations, but rather on geometric projections.
While most existing tools focus on the  $\ell_1$ and $\ell_\infty$ norms,
we provide an efficient, highly parallel implementation to certify $\ell_2$-robustness.
Our implementation outperforms existing exact tools by multiple orders of magnitude, while empirically maintaining the same or better precision under a time constraint.
Additionally, we show that our approach is particularly suited to scale up network verification to \emph{deeper} networks---a promising step towards verifying large, state-of-the-art models.

%Interesting directions for future work would be to extend our implementation to support convolutional layers, to efficiently compute projections for other $\ell_p$ norms, and to explore new search heuristics to detect adversarial examples faster when they exist.

\paragraph{Acknowledgments.}
The work described in this paper has been supported by the Software Engineering Institute under its FFRDC Contract No. FA8702-15-D-0002 with the U.S. Department of Defense, Bosch Corporation, an NVIDIA GPU grant, NSF Award CNS-1801391, DARPA GARD Contract HR00112020006, a Google Faculty Fellowship, and the Alfred P. Sloan Foundation.

\bibliography{mybibliography}
\bibliographystyle{iclr2021_conference}

\newpage

\appendix

% !TEX root=./paper.tex

\section{Correctness Proofs}
\label{sec:proofs}

\subsection{Proof of Theorem~\ref{thm:correctness}}\label{sec:fgp_proof}

We show that when FGP returns \notrobust, there exists an adversarial example, and when it returns \robust, the model is locally robust at $x$, with radius $\epsilon$. However, the algorithm may also return \unknown, in which case we do not claim anything about the robustness of the model.

%\begin{theorem}\label{thm:correctness}
%(1) When FGP returns \notrobust, there exists an adversarial example, $p$, such that $\norm{x - p} \leq \epsilon$ and $\modelclass(x) \neq \modelclass(p)$. 
%(2) When FGP returns \robust, $\forall x' . \|x-x'\|_p \le \epsilon \Longrightarrow F(x) = F(x')$.
%\end{theorem}

\begin{proof}
In the first case, where FGP returns \notrobust, the proof of Theorem~\ref{thm:correctness} is trivial:
we exhibit a point, $p$, such that $\|x-p\| \leq \epsilon$, and for which $F(x) \neq F(p)$.
%We exhibit a point, $p$, that is the projection of $x$ onto a decision boundary constraint, $c$, that is at distance less than $\epsilon$ from $x$ for which $m(x) \neq m(p)$.

The interesting case is when FGP returns \robust.
We prove by contradiction that in this case, $x$ is in fact locally robust with radius $\epsilon$.

Let us assume for the sake of contradiction that FGP returns \robust, but there exists a point, $p$, such that $\norm{x - p} \leq \epsilon$ and $F(x) \neq F(p)$.
Let $\ar_x$ and $\ar_p$ be the activation regions associated with $x$ and $p$ respectively.

We define a \emph{path} of activation regions as a sequence, $\ar_0, \dots, \ar_k$, of activation regions such that the underlying activation patterns $\ap_i$ and $\ap_{i+1}$ differ in exactly one neuron for all $i$, and there exists at least one input, $x_i$, that has activation pattern $\ap_i$ for all $i$.
For instance, in a network with three neurons, if $\ap_0 = (true, true, false)$, $\ap_1 = (true, false, false)$, and $\ap_2 = (true, false, true)$, and  there exist inputs, $x_0$, $x_1$, and $x_2$ with activation patterns $\ap_0$, $\ap_1$, and $\ap_2$, then $\langle\ar_0$, $\ar_1$, $\ar_2\rangle$ is a path.

Our proof relies on three facts, that we prove hereafter:
\begin{enumerate}
\item \label{proof_point_path}
  There exists a path, $P$, from $\ar_x$ to $\ar_p$ where each region in the path contains at least one input at distance less than $\epsilon$ from $x$, and either $\ar_x = \ar_p$ or all $\ar_i, \ar_j$ in the path are different.
\item \label{proof_point_visited}
  Our algorithm visits all regions in the path, $P$.
\item \label{proof_point_return}
  If a visited activation region contains an adversarial input, our algorithm either detects it, returning \notrobust, or returns \unknown.
\end{enumerate}

Together, \emph{(\ref{proof_point_path})}, \emph{(\ref{proof_point_visited})}, and \emph{(\ref{proof_point_return})} imply that if an adversarial point, $p$, exists, it resides in an activation region that would have been checked by FGP, which would have resulted in the algorithm returning \notrobust or \unknown, contradicting the assumption that it returned \robust.

%\todo{perhaps the proof should simply state that there is a finite number of neurons, there are no loops, so  there is a finite number of activations
% and the algorithm, in a naive version visits all of them; is the impl less naive?}
%\todo{Bryan: That might be a bit too telegraphed for the reviewers.  Plus a formal proof makes the paper seem even more serious!}

\paragraph{(\ref{proof_point_path}) Existence of a Path}
Consider the segment going from $x$ to $p$ in a straight line.
As $\|x - p\| \leq \epsilon$, all points on this segment are also at distance less than $\epsilon$.
% As $m$ is a neural network with continuous, piecewise linear activations, $m$ is a continuous function, as are the activation functions, $m_n$ for each of its internal neurons, $n$.
As $f$ is a neural network with ReLU activations, $f$ is a continuous function, as are the activation functions, $f_u$ for each of its internal neurons, $u$.
% Furthermore, input points at the boundary between two activation regions (e.g., for ReLU networks, points, $z$, such that $m_n(z) = 0$) belong to \emph{both} activation regions.
Furthermore, input points at the boundary between two activation regions (i.e., $z$, such that $f_u(z) = 0$) belong to \emph{both} activation regions.
Therefore, listing all activation regions on the segment between $x$ and $p$ yields a path, $P$, from $\ar_x$ to $\ar_p$, with each region on the path containing an input point, $x_i$, on the segment, such that $\|x - x_i\| \leq \epsilon$.

That each $\ar_i$, $\ar_j$ in $P$ is unique follows from the convexity of activation regions:
If there exist $\ar_i = \ar_j$ in $P$ such that there exists an $\ar_k$ in between, then $\ar_i = \ar_j$ cannot be convex, as there exists a line segment with its end points in $\ar_i$ and $\ar_j$ that is not entirely contained within the region $\ar_i = \ar_j$.
This ensures that paths are of finite length.

\paragraph{(\ref{proof_point_visited}) Exploration by the Algorithm}
Given the existence of the path, $P$, from $\ar_x$ to $\ar_p$, we now prove that FGP would visit all activation regions in $P$ if it returns \robust.
We proceed by induction on the length of paths induced (similarly to as above) by a line segment included in the $\epsilon$-ball centered on $x$.

In the base case, if the path is of length one, then it contains only $\ar_x$, and the claim holds trivially since FGP starts by visiting $\ar_x$.

In the inductive case, let us assume that for any path with length at most $k$ induced by a segment, $s$, beginning at $x$ with $\norm{s} \leq \epsilon$, FGP visits all regions in the path.
Now consider a path, $P' = \ar_0, \dots, \ar_k$, of length $k+1$, induced by a segment, $s'$, beginning at $x$ with $\norm{s'} \leq \epsilon$.
Since $\ar_{k-1}$ is on the path, there exists a point, $x_{k-1}$, on $s'$ such that the sub-segment from $x$ to $x_{k-1}$ induces a path of length $k$; thus we can apply our induction hypothesis to conclude that FGP visits $\ar_0, \dots, \ar_{k-1}$.
Now, since $\ar_{k-1}$ and $\ar_k$ are neighbors in $P'$, they must share some boundary, $C$, that is intersected by $s'$.
Thus, since $\norm{s'} \leq \epsilon$, $d(x, C) \leq \epsilon$; thus when FGP visits $\ar_{k-1}$, it will add $\ar_k$ to the queue via $C$.
Therefore, since FGP returns \robust only when all regions in the queue have been visited, FGP will visit $\ar_{k}$, concluding the proof of (\ref{proof_point_visited}).

\paragraph{(\ref{proof_point_return}) Detection of Adversarial Examples}
We conclude by proving that if there exists an adversarial example in a region visited by FGP, then we either return \notrobust or \unknown.

If $p$ in $\ar_p$ is an adversarial example, then $F(x) \neq F(p)$.
By continuity of $f$, this means that there exists an input, $p'$ on the segment from $x$ to $p$ that is exactly on a decision boundary, $C$.
As $\norm{x - p'} \leq \norm{x - p} \leq \epsilon$, $C$ must have been analyzed when exploring $\ar_p$.

However, when analyzing a decision boundary, FGP always returns either \notrobust or \unknown.

Thus, the decision boundary in the activation region containing $p$ would have been analyzed by FGP; this yields a contradiction, as the algorithm must have returned \notrobust or \unknown rather than \robust.
\end{proof}

%\subsection{Proof of Theorem~\ref{thm:correctness} for Tree-Based Exploration}\label{sec:tree_proof}
%
%\input{proof_tree}

\subsection{Proof of Loop Invariant~\ref{thm:loopinvariant}}
\label{sec:loop_proof}

The proof of lower-bound variant is similar to the proof of FGP in Section~\ref{sec:fgp_proof}.
It relies on the following loop invariant.

%\begin{loopinvariant}\label{thm:loopinvariant}
%(1) At each iteration, all activation regions at distance less than $bound$ from $x$ were previously visited, (2) $bound$ is always smaller than $\epsilon$,
%and (3) there is no adversarial point at distance less than $bound$.
%\end{loopinvariant}
\begin{proof}
Loop Invariant~\ref{thm:loopinvariant} trivially holds on entry of the loop since $bound$ is initialized to 0.
Proving that the invariant is maintained is more interesting. 
Suppose that $bound$ increases to $bound'$. 
It must then be shown that there is no unvisited activation region at distance less than $bound'$.
We proceed again by contradiction:
assume there was such a region, $\ar$, containing a point, $p$, such that $\norm{x - p} \leq bound'$.
Again, let us consider the segment from $x$ to $p$, and the path, $P$, it induces.
Let us consider $\ar_i$, the first region of $P$ at distance greater than $bound$ that was not previously visited.
If no such region exists, then $\ar$ is at distance less than $bound$ from $x$, and so by our induction hypothesis, it was already visited.
% and therefore visited since the queue is a priority queue and $bound < bound'$.
otherwise, $\ar_{i-1}$ was visited, and the activation constraint, $C$, between $\ar_{i-1}$ and $\ar_i$ is such that $d(x, C) \leq bound' \leq \epsilon$.
Therefore, $C$ (which leads to $\ar_i$) was already added to the queue with priority less than $bound'$, and by virtue of the priority queue, it was explored before the current iteration, which yields a contradiction.

If $bound$ does not increase, the invariant still trivially holds.
This case can happen because our computation of the distance to constraints is an underapproximation of the true distance to the feasible portion of the constraint.
%, as is illustrated in Figure~\ref{fig:boundconstraint}.
\end{proof}

\section{Computing Activation Constraints And Decision Boundaries}
\label{sec:appendix:constraints}

Recall that an activation constraint, $C_u$, which is satisfied when $\modelfn_u(x) \geq 0 \Longleftrightarrow \ap(u)$,
is a linear constraint with coefficients $w_u$, and intercepts, $b_u$.
The computation of these weights and intercepts does not depend on a particular point in $\ar$ --- only on the activation pattern, $\ap$.
Thus, \emph{we can compute the boundaries of an activation region, $\ar$, knowing only the corresponding activation pattern, $\ap$}.

In practice, the coefficients, $w_u$, correspond to the gradient of $\modelfn_u$ with respect to its inputs, evaluated at a point in $\ar$.
However, frameworks that perform automatic differentiation typically require a concrete point for evaluation of the gradient.
Thus, we compute the gradient via backpropagation with the activation vectors, $a_i$, where the position in $a_i$ corresponding
to neuron, $u$, takes value 1 if $\ap(u)$ is $true$ and 0 otherwise.
The intercepts, $b_u$, are computed via a forward computation using the activation vectors, with $x$ set to 0.
These operations can easily be implemented to run efficiently on a GPU.

Decision boundaries in an activation region, $\ar$, are computed similarly to internal activation constraints;
we take the linearization of the network in $\ar$. Since the network is piecewise linear, and linear within each activation region, this linearization is exact for the region being considered.

\section{Hyperparameters}
\label{sec:appendix:hyper}

Here, we provide details on the hyperparameters used to train the models used in our evaluation and to conduct the PGD~\citep{madry2018towards} attacks used to obtain the upper bounds on the verified robust accuracy (VRA).

\paragraph{CNN Architecture.}
We used a CNN architecture that has been used for benchmarking verification by \citep{wong2017provable} and \citep{croce19mmr}.
It contains 2 convolutional layers, each using $4\times4$ filters with a $2\times2$ stride, with 16 and 32 channels respectively, followed by a dense layer of 100 nodes.

\paragraph{PGD Adversarially-trained Models.}
To train models for robustness, we used PGD adversarial training~\citep{madry2018towards}. 
For training, we used the $\ell_2$ norm, and let $\epsilon = 2.5$---10 times the $\epsilon$ we verify with in order to have a higher fraction of verifiably-robust points.
We trained each model for 20 epochs with a batch size of 128 and 50 PGD steps.

\paragraph{Maximum-Margin-Regularization-trained Models.}
To train models for both robustness and verifiability, we used maximum margin regularization~\citep{croce19mmr} (MMR).
MMR has several hyperparameters including $\gamma_B$ and $\gamma_D$, which correspond to the desired distance of the training points from the internal boundaries and decision boundaries respectively (these can be seen as acting similarly to the choice of $\epsilon$ in PGD training); $n_B$ and $n_D$, which specify the number of internal and decision boundaries to be moved in each update; and $\lambda$, which specifies the relative weight of the regularization as opposed to the regular loss.
See \citep{croce19mmr} for more details on these hyperparameters.
We set $\gamma_B = \gamma_D = 2.5$, $n_B = 100$, $n_D = 9$, and $\lambda = 0.5$, and trained each model for 20 epochs with a batch size of 128.

\paragraph{ReLU-Stability-trained Models.}
We also used ReLU stability (RS)~\citep{madry19relu} to regularize models for verifiable robustness.
We trained using RS with PGD adversarial loss and $\ell_1$ weight regularization, as was done by \citeauthor{madry19relu}.
We weighted the RS loss by $\alpha = 2.0$, using an $\epsilon$ (i.e., the distance over which the ReLU activations should remain stable) of $8/255$, and weighted the PGD adversarial loss by $\beta = 1.0$, using an $\epsilon$ (i.e., the target robustness radius) of $36/255$.
We scheduled the $\ell_1$ regularization to decay from $10^{-2}$ to $10^{-3}$ over the course of training, and trained for 100 epochs with a batch size of 128.

\paragraph{PGD Attacks.}
We obtained an upper bound on the VRA by performing PGD attacks on all the correctly-classified points for which every method either timed out or reported \unknown;
the upper bound is the best VRA plus the fraction of points that were correctly-classified, undecided, and for which the PGD attack did not successfully find an adversarial example.
We conducted these attacks with an $\ell_2$ bound of $\epsilon = 0.25$ with 1,000 PGD steps.

\paragraph{Arbitrary Test Instances}
To arbitrarily pick the 100 test instances, we used the \lstinline{random.randint} function from Numpy, setting the initial seed to 4444.
The indices of the test instances are 8361, 7471, 8091, 4275, 5205, 6886, 5341, 4026, 4554, 5835, 8181,
       6124, 6309, 1875, 7910, 6018, 1985, 6912, 8676, 2376, 9143, 5527,
       5540, 4661, 1910, 7498, 2190,  789, 1672, 3393, 1925,  841, 8333,
       4644, 5648,  138, 7872, 5204,  441, 3951, 3812, 3983, 8598, 8334,
       5666,    5, 1014, 9148, 5993, 2182, 4141, 5385, 2774, 9927, 7507,
       9097, 7047, 7319, 1638, 9535, 8889, 1196, 4992, 9223, 6525, 9577,
       2266, 1748, 6462, 2969, 4866,  271, 9890, 2814, 9848, 8513,  289,
       9002, 1103,  506, 1576, 1035, 9808, 8059, 2424, 6072, 6577, 1463,
       5573, 8995,  100, 1457, 8103, 7835, 2491, 9481, 8053, 2478, 8132,
       7033.

\newpage

\section{Breakdown of Run-times for Decided Points}

\begin{figure}[h]
\centering
\begin{subfigure}[t]{\textwidth}
  \centering
  \includegraphics[page=1, width=\textwidth]{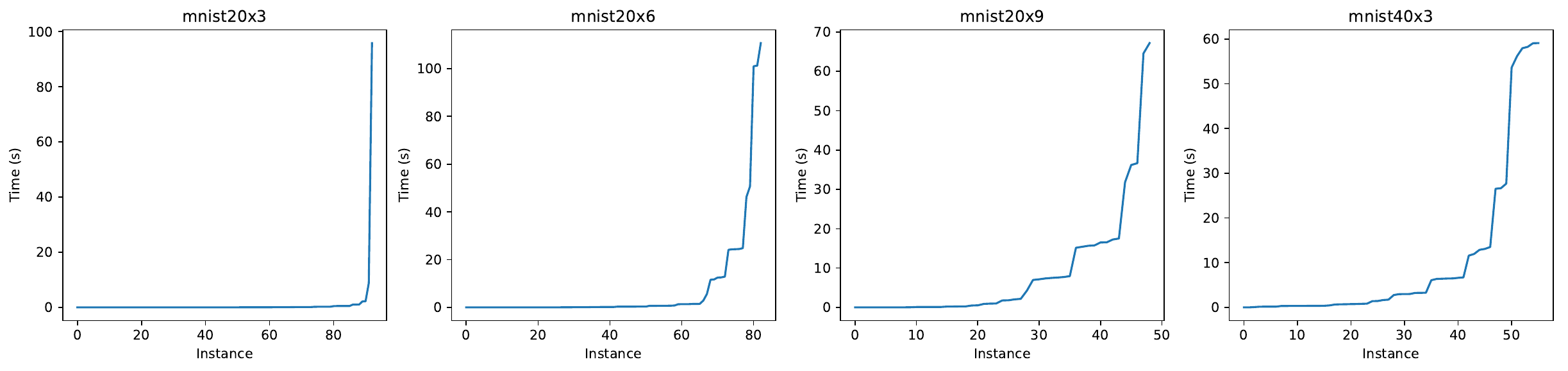}%
  \caption{}
  \label{fig:cactus}
\end{subfigure}
\begin{subfigure}[t]{\textwidth}
  \centering
  \includegraphics[page=1, width=\textwidth]{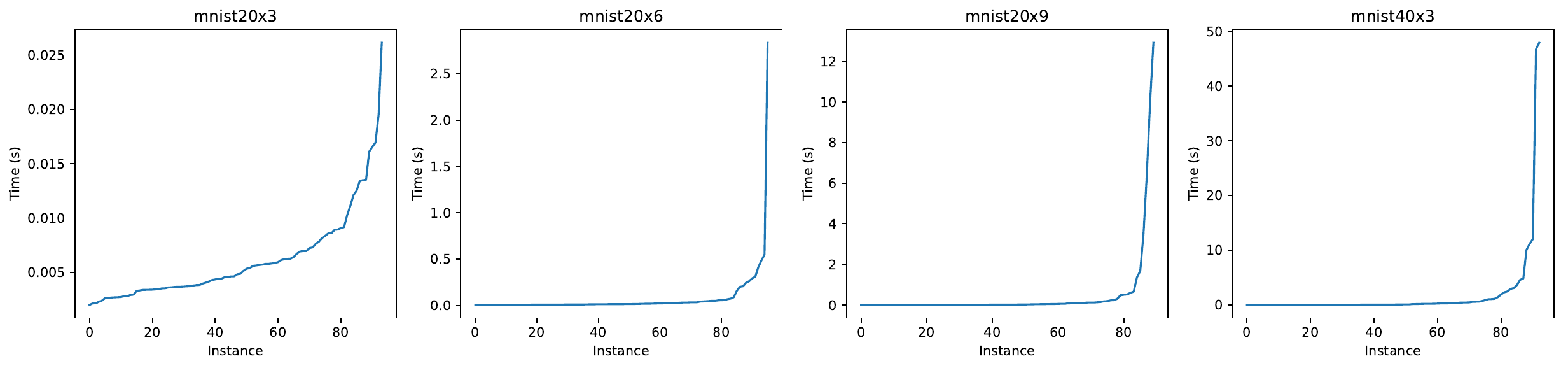}%
  \caption{}
  \label{fig:cactus_linf}
\end{subfigure}%
\caption{
  Cactus plots showing the time taken for each instance that is decided as either \robust or \notrobust. The instances are ordered by increasing time-per-instance. Results are shown for the PGD adversarially-trained models for both the $\ell_2$ norm (\subref{fig:cactus}) and the $\ell_\infty$ norm (\subref{fig:cactus_linf}).
}
\end{figure}

\end{document}